\documentclass[11pt,a4paper]{article}
\usepackage[hyperref]{acl2019}
\usepackage{times}
\usepackage{latexsym}

\usepackage{url}

\usepackage{paralist}
\usepackage{xspace}
\usepackage{mathrsfs, amsmath, amssymb}
\usepackage{graphicx}
\usepackage{subcaption}
\usepackage[]{xcolor}

\usepackage{booktabs}
\usepackage[tight-spacing=true]{siunitx}
\usepackage{todonotes}
\usepackage{etoolbox,multirow,adjustbox}
\robustify\bfseries

\newcommand{\vect}[1]{\ensuremath{\mathbf{#1}}}
\newcommand\xvect{\vect{x}}

\newcommand\zvect{\vect{z}}
\newcommand\hvect{\vect{h}}

\newcommand\expe{\ensuremath{\mathop{\mathbb{E}}}}
\newcommand\dist{\ensuremath{\mathcal{D}}}

\newcommand{\ex}[1]{\textit{#1}\xspace}
\newcommand{\scnn}{\resource{s-cnn}}
\newcommand{\bcnn}{\resource{m-cnn}}
\newcommand{\gen}{\resource{gen}}
\newcommand{\dsda}{\resource{dsda}}
\newcommand{\csda}{\resource{csda}}

\newcommand{\mathcalA}{\ensuremath{\mathcal{F}}}
\newcommand{\mathcalC}{\ensuremath{\mathcal{Y}}}

\newcommand{\resource}[1]{\textsc{#1}\xspace}
\newcommand{\D}[1]{\ensuremath{\operatorname{#1}\xspace}}
\newcommand{\MLP}{\ensuremath{\operatorname{MLP}\xspace}}
\newcommand{\f}{\ensuremath{\operatorname{f}\xspace}}
\newcommand{\CNN}{\ensuremath{\operatorname{CNN}\xspace}}

\newcommand{\Dirichlet}{\ensuremath{\operatorname{Dirichlet}\xspace}}

\newcommand{\BetaD}{\ensuremath{\operatorname{Beta}\xspace}}

\newcommand\tabref[1]{Table~\ref{#1}}
\newcommand\figref[1]{Figure~\ref{#1}}
\newcommand\secref[1]{Section~\ref{#1}}
\newcommand\appref[1]{Appendix~\ref{#1}}

\usepackage[utf8]{inputenc}
\usepackage{tikz,pgfplots}
\usetikzlibrary{positioning}
\usetikzlibrary{calc}
\usetikzlibrary{fit}
\usetikzlibrary{decorations.text}
\usetikzlibrary{decorations.pathreplacing}
\usetikzlibrary{shapes.misc}
\usetikzlibrary{shapes.geometric,arrows}

\definecolor{color0}{rgb}{0.917647058823529,0.917647058823529,0.949019607843137}
\definecolor{color1}{rgb}{0.298039215686275,0.447058823529412,0.690196078431373}
\definecolor{color2}{rgb}{0.866666666666667,0.517647058823529,0.32156862745098}
\definecolor{color3}{rgb}{0.333333333333333,0.658823529411765,0.407843137254902}
\definecolor{color4}{rgb}{0.768627450980392,0.305882352941176,0.32156862745098}
\definecolor{color5}{rgb}{0.505882352941176,0.447058823529412,0.701960784313725}

\def\Bcolour{7e1e9c}
\def\Dcolour{15b01a}
\def\Ecolour{0343df}
\def\Kcolour{ff81c0}

\aclfinalcopy 
\def\aclpaperid{1984} 

\title{Semi-supervised Stochastic Multi-Domain Learning using Variational Inference}

\author{Yitong Li \\ \And
  Timothy Baldwin \\
  School of Computing and Information Systems \\
  The University of Melbourne, Australia \\
  {\small \tt yitongl4@student.unimelb.edu.au} \\
  {\small \tt \{tbaldwin,tcohn\}@unimelb.edu.au} \\  \And
  Trevor Cohn \\
 }

\date{}

\begin{document}
\maketitle
\begin{abstract}
Supervised models of NLP rely on large collections of text which closely resemble the intended testing setting. 
Unfortunately matching text is often not available in sufficient quantity, and moreover, within any domain of text, data is often highly heterogenous.
In this paper we propose a method to distill the important domain signal as part of a multi-domain learning system, using a latent variable model in which parts of a neural model are stochastically gated based on the inferred domain.
We compare the use of discrete versus continuous latent variables, operating in a domain-supervised or a domain semi-supervised setting, where the domain is known only for a subset of training inputs.
We show that our model leads to substantial performance improvements over competitive benchmark domain adaptation methods, including methods using adversarial learning.
\end{abstract}

\section{Introduction}

Text corpora are often collated from several different sources, such as news, literature, micro-blogs, and web crawls, raising the problem of learning NLP systems from heterogenous data, and how well such models transfer to testing settings.
Learning from these corpora requires models which can generalise to different domains, a problem known as transfer learning or domain adaptation \cite{blitzer2007biographies,daume2007frustratingly,Joshi2012multi,C16-1038}.
In most state-of-the-art frameworks, the model has full knowledge of the domain of instances in the training data, and the domain is treated as a discrete indicator variable.
However, in reality, data is often messy, with domain labels not always available, or providing limited information about the style and genre of text. 
For example, web-crawled corpora are comprised of all manner of text, such as news, marketing, blogs, novels, and recipes, however the type of each document is typically not explicitly specified.
Moreover, even corpora that are labelled with a specific domain might themselves be instances of a much more specific area, e.g., ``news'' articles will cover politics, sports, travel, opinion, etc.
Modelling these types of data accurately requires knowledge of the specific domain of each instance, as well as the domain of each test instance, which is particularly problematic for test data from previously unseen domains.

A simple strategy for domain learning is to jointly learn over all the data with a single model, where the model is not conditioned on domain, and directly maximises $p(y|\xvect)$, where $\xvect$ is the text input, and $y$ the output (e.g.\ classification label).
Improvements reported in multi-domain learning \cite{daume2007frustratingly,C16-1038} have often focused on learning twin representations (\emph{shared} and \emph{private} representations) for each instance.
The private representation is modelled by introducing a domain-specific channel conditional on the domain, and the shared one is learned through domain-general channels.
To learn more robust domain-general and domain-specific channels, adversarial supervision can be applied in the form of either domain-conditional or domain-generative methods \cite{DBLP:conf/emnlp/LiuQH16,yitong2018naacl}.

Inspired by these works, we develop a method for the setting where the domain is unobserved or partially observed, which we refer to as unsupervised and semi-supervised, respectively, with respect to domain.
This has the added benefit of affording robustness where the test data is drawn from an unseen domain, through modelling each test instance as a mixture of domains.
In this paper, we propose methods which use latent variables to characterise the domain, by modelling the discriminative learning problem $p(y|\xvect) = \int_z p(z|\xvect)p(y|\xvect,z)$, where $z$ encodes the domain, which must be marginalised out when the domain is unobserved.
We propose a sequence of models of increasing complexity in the modelling of the treatment of $z$, ranging from a discrete mixture model, to a continuous vector-valued latent variable (analogous to a topic model; \citet{blei2003latent}), modelled using Beta or Dirichlet distributions.
We show how these models can be trained efficiently, using either direct gradient-based methods or variational inference  \cite{DBLP:conf/nips/KingmaMRW14}, for the respective model types.
The variational method can be applied to domain and/or label semi-supervised settings, where not all components of the training data are fully observed.

We evaluate our approach using sentiment analysis over multi-domain product review data and 7 language identification benchmarks from different domains, showing that in out-of-domain evaluation, our methods substantially improve over benchmark methods, including adversarially-trained domain adaptation \cite{yitong2018naacl}.
We show that including additional domain unlabelled data gives a substantial boost to performance, resulting in transfer models that often outperform domain-trained models, to the best of our knowledge, setting a new state of the art for the dataset.

\section{Stochastic Domain Adaptation}

In this section, we describe our proposed approaches to Stochastic
Domain Adaptation (SDA), which use latent variables to represent an implicit `domain'.
This is formulated as a joint model of output classification label, $y$ and latent domain $z$, which are both conditional on $\xvect$,
\[ p(y, z|\xvect)  = p_\phi(\vect{z}|\xvect)p_\theta(y|\xvect,\vect{z})  \, . \]
The two components are the prior, $p_\phi(\vect{z}|\xvect)$, and classifier likelihood, $p_\theta(y|\xvect,\vect{z})$, 
which are parameterised by $\phi$ and $\theta$, respectively.
We propose  several different choices of prior, based on the nature of $\vect{z}$, that is, whether it is: (i) a discrete value (``\dsda'', see \secref{sec:discrete}); or (ii) a continuous vector, in which case we experiment with different distributions to model $p(\vect{z}|\xvect)$ (``\csda'', see \secref{sec:continuous}).

\subsection{Stochastic Channel Gating} \label{sec:gating}
 
For all of our models the likelihood,  $p_\theta(y|\xvect,\vect{z}) $, is formulated as a multi-channel neural model, where $\vect{z}$ is used as a gate to select which channels should be used in representing the input.
The model comprises $k$ channels, with each channel computing an independent hidden representation,
\[ \vect{h}_{i} = \D{CNN}_{i} (\vect{x} ; \theta) \vert_{i=1}^k \] 
using a convolutional neural network.\footnote{Our approach is general, and could be easily combined with other methods besides CNNs.}
The value of $z$ is then used to select the channel, by computing
$ \vect{h} = \sum_{i=1}^k z_k \vect{h}_{i} $, 
where we assume $\vect{z} \in \mathbb{R}^k$ is a continuous vector. 
For the discrete setting, we represent integer $z$ by its 1-hot encoding $\vect{z}$, 
in which case 
$\vect{h} = \vect{h}_{z}$.
The final step of the likelihood passes $\vect{h}$ through a MLP with a single hidden layer, followed by a softmax, which is used to predict class label $y$.

\begin{figure*}[t!]
\begin{subfigure}[b]{ \columnwidth }
  \begin{tikzpicture}[line width=0.02cm]

    \node[align=center,minimum size=0.3cm] at (-3.3, 0) {$\xvect$};

    \node[align=center,minimum size=0.5cm] at (-1.3, 0.2) (B1) {\small $\CNN_k(\theta_k)$};
    \node[draw=none,align=center,minimum size=0.5cm] at (-1.3, 1.2) {\small $\CNN_1(\theta_1)$};
    \node[draw=none,align=center,minimum size=0.5cm] at (-1.3, 0.7) {\small $\cdots$};

    \filldraw[fill=green!20!white, draw=green!20!white,rounded corners] (-0.3, -0.3)rectangle(0.3, 0.3);
    \node[draw=none,align=center,minimum size=0.5cm] at (0.0, -0.0) {\small $\hvect_k$};
    \filldraw[fill=green!20!white, draw=green!20!white,rounded corners] (-0.3, +0.7)rectangle(0.3, 1.3);
    \node[draw=none,align=center,minimum size=0.5cm] at (0.0, +1.0) {\small $\hvect_1$};
    \node[draw=none,align=center,minimum size=0.5cm] at (0.0, 0.5) {\small $\cdots$};
    \draw[draw=green!80!white,rounded corners] (-0.4, -0.4)rectangle(0.4, 1.4);

    \draw[->,rounded corners] (-2.8, 0.0)--(-0.35, 0.0);
    \draw[->,rounded corners] (-2.8, 0.0)--(-2.5, 0.0)--(-2.5,1.0)--(-0.35,1.0);

    \draw[draw=green!80!white,->,rounded corners] ( 0.4, 0.5)--( 1.0, 0.5);

    \node[draw=none,align=center,text width=6cm] at (1.3, 0.5) {$\bigodot$};

    \draw[draw=green!80!white,->,rounded corners] ( 1.6, 0.5)--( 2.0, 0.5);

    \node[draw=none,align=center,text width=6cm] at (2.3,+0.5) {$\hvect$};

    \draw[->,rounded corners] ( 2.5, 0.5)--( 3.0, 0.5);

    \node[draw=none,align=center,text width=6cm] at (3.2,+0.5) {$y$};

    \node[draw=none,align=center,minimum size=0.5cm] at (-1.4, 2.2) {\small $\CNN(\phi)$};

    \filldraw[fill=white, draw=blue!80!white] ( 0.0, 2.0 ) circle (8pt);
    \node[draw=none,align=center,minimum size=0.5cm] at (0.0, 2.0) {\small $p$};


    \filldraw[fill=blue!20!white, draw=blue!80!white,rounded corners] (1.0, 1.7)rectangle(1.6, 2.3);
    \node[draw=none,align=center,minimum size=0.5cm] at (1.3, +2.0) {\small ${\vect{z}}$};

    \draw[draw=blue!80!white,->,rounded corners] (-3.3, 0.3)--(-3.3, 2.0)--(-0.35, 2.0);
    \draw[draw=blue!80!white,->,rounded corners] ( 0.35, 2.0)--( 0.95, 2.0);

    \draw[draw=blue!80!white,->,rounded corners] (1.3, 1.7)--( 1.3, 0.75);







    

\end{tikzpicture}

%
\caption{\dsda}
\label{fig:dsda}
\end{subfigure}
~
\begin{subfigure}[b]{\columnwidth}
  \begin{tikzpicture}[line width=0.02cm]

    \node[align=center,minimum size=0.3cm] at (-3.3, 0) {$\xvect$};

    \node[align=center,minimum size=0.5cm] at (-1.3, 0.2) (B1) {\small $\CNN_k(\theta_k)$};
    \node[draw=none,align=center,minimum size=0.5cm] at (-1.3, 1.2) {\small $\CNN_1(\theta_1)$};
    \node[draw=none,align=center,minimum size=0.5cm] at (-1.3, 0.7) {\small $\cdots$};

    \filldraw[fill=green!20!white, draw=green!20!white,rounded corners] (-0.3, -0.3)rectangle(0.3, 0.3);
    \node[draw=none,align=center,minimum size=0.5cm] at (0.0, -0.0) {\small $\hvect_k$};
    \filldraw[fill=green!20!white, draw=green!20!white,rounded corners] (-0.3, +0.7)rectangle(0.3, 1.3);
    \node[draw=none,align=center,minimum size=0.5cm] at (0.0, +1.0) {\small $\hvect_1$};
    \node[draw=none,align=center,minimum size=0.5cm] at (0.0, 0.5) {\small $\cdots$};
    \draw[draw=green!80!white,rounded corners] (-0.4, -0.4)rectangle(0.4, 1.4);

    \draw[->,rounded corners] (-2.8, 0.0)--(-0.35, 0.0);
    \draw[->,rounded corners] (-2.8, 0.0)--(-2.5, 0.0)--(-2.5,1.0)--(-0.35,1.0);

    \draw[draw=green!80!white,->,rounded corners] ( 0.4, 0.5)--( 1.0, 0.5);

    \node[draw=none,align=center,text width=6cm] at (1.3, 0.5) {$\bigodot$};

    \draw[draw=green!80!white,->,rounded corners] ( 1.6, 0.5)--( 2.0, 0.5);

    \node[draw=none,align=center,text width=6cm] at (2.3,+0.5) {$\hvect$};

    \draw[->,rounded corners] ( 2.5, 0.5)--( 3.0, 0.5);

    \node[draw=none,align=center,text width=6cm] at (3.2,+0.5) {$y$};

    \node[align=center,minimum size=0.5cm] at (-1.4, 3.2) (B1) {\small $\CNN(\sigma)$};
    \node[draw=none,align=center,minimum size=0.5cm] at (-1.4, 2.2) {\small $\CNN(\phi)$};

    \filldraw[fill=white, draw=blue!80!white] ( 0.0, 2.0 ) circle (8pt);
    \node[draw=none,align=center,minimum size=0.5cm] at (0.0, 2.0) {\small $p$};
    \filldraw[fill=white, draw=orange!80!white] ( 0.0, 3.0 ) circle (8pt);
    \node[draw=none,align=center,minimum size=0.5cm] at (0.0, +3.0) {\small $q$};

    \node[draw=none,align=center,minimum size=0.5cm] at (0.7, +2.5) {\small $\sim$};

    \filldraw[fill=blue!20!white, draw=blue!80!white,rounded corners] (1.0, 2.2)rectangle(1.6, 2.8);
    \node[draw=none,align=center,minimum size=0.5cm] at (1.3, +2.5) {\small $\widehat{\zvect}$};

    \draw[draw=blue!80!white,->,rounded corners] (-3.25, 0.3)--(-3.25, 2.0)--(-0.35, 2.0);
    \draw[draw=orange!80!white,->,rounded corners] (-3.35, 0.3)--(-3.35, 3.0)--(-0.35,3.0);

    \draw[draw=blue!80!white,->,rounded corners] (1.3, 2.2)--( 1.3, 0.75);

    \node[draw=none,align=center,minimum size=0.5cm] at (0.0, 4.0) {$y, d$};
    \draw[draw=orange!80!white,->,rounded corners] (0.0, 3.8)--( 0.0, 3.35);

    \draw[draw=red!80!white,dashed,rounded corners] (-0.4, 1.6)rectangle(0.4, 3.4);
    \draw[draw=red!80!white,dashed,->,rounded corners] (0.4, 3.2)--(0.6, 3.4)--( 2.6, 3.4);
    \node[draw=none,align=center,minimum size=0.5cm] at (3.0, 3.4) {$\dist_{KL}$};





    

\end{tikzpicture}

%
\caption{\csda}
\label{fig:mod}
\end{subfigure}
\caption{Model architectures for latent variable models, \dsda and \csda, which differ in the treatment of the latent variable, which is discrete ($d \in [1,k]$), or a continuous vector ($\hat{\mathbf{z}} \in \mathbb{R}^k$). The lower green model components show $k$ independent convolutional network components, and the blue and yellow component the prior, $p$, and the variational approximation, $q$, respectfully. The latent variable is used to gate the $k$ hidden representations (shown as $\bigodot$), which are then used in a linear function to predict a classification label, $y$. During training \csda draws samples ($\sim$) from $q$, while during inference, samples are drawn from $p$.}
\label{fig:dsda_and_csda}
\end{figure*}
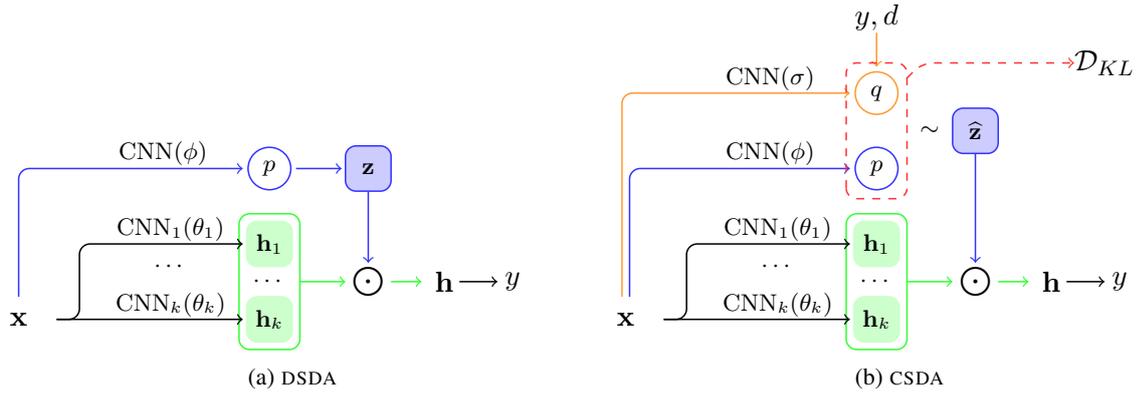

\subsection{Discrete Domain Identifiers} \label{sec:discrete} We now
turn to the central part of our method, the prior component.  The
simplest approach, \dsda (see \figref{fig:dsda}), uses a discrete latent
variable, i.e., $z \in [1, k]$ is an integer-valued random variable, and
consequently the model can be considered as a form of mixture model.
This prior predicts $z$ given input $\xvect$, which is modelled using a
neural network with a softmax output.  Given $z$, the process of
generating $y$ is as described above in \secref{sec:gating}.  The
discrete model can be trained for the maximum likelihood estimate using
the objective,
\begin{align} \label{eq:dsdaloss}
\log p(y|\xvect) = \log\sum_{z=1}^{k} p_\phi(z|\xvect) p_\theta(y|\xvect, z),
\end{align}
which can be computed tractably,%
\footnote{This arises from the finite summation in (\ref{eq:dsdaloss}), which requires each of the $k$  components to be computed separately, and their results summed. This procedure permits standard gradient back-propagation. }
and scales linearly in $k$.

\dsda can be  applied with supervised or semi-supervised domains, by maximising the likelihood $p(z=d|\xvect)$ when the ground truth domain $d$ is observed.  We refer to this setting as ``\dsda+sup.\@'' or ``\dsda+semisup'', respectively, noting that in this setting we assume the number of channels, $k$, is equal to the known inventory of domains, $D$.

\subsection{Continuous Domain Identifiers} \label{sec:continuous}
For the \dsda model to work well requires sufficiently large $k$, such that all the different types of data can be clearly separated into individual mixture components.
When there is not a clear delineation between domains, the inferred domain posterior is likely to be uncertain, and the approach reduces to an ensemble technique.
Thus, we introduce the second modelling approach as Continuous domain identifiers (\csda), inspired by the way in which LDA models the documents as mixtures of several topics~\cite{blei2003latent}. 

A more statistically efficient method would be to use binary functions as domain specifiers, i.e., $\vect{z} \in \{ 0,1\}^k$, effectively allowing for exponentially many domain combinations ($2^k$).
Each element of the domain $z_i$ acts as a gate, or equivalently, attention, governing whether hidden state $\hvect_i$ is incorporated into the predictive model.
In this way, individual components of the model can specialise to a very specific topic such as politics or sport, and yet domains are still able to combine both to produce specialised representations, such as the politics of sport.
The use of a latent bit-vector renders inference intractable, due to the marginalisation over exponentially many states.
For this reason, we instead make a continuous relaxation, such that $\vect{z} \in \mathbb{R}^k$ with each scalar $z_i$ being drawn from a probability distribution parameterised as a function of the input $\xvect$.
These functions can learn to relate aspects of $\xvect$ with certain domain indexes, e.g., the use of specific words like \ex{baseball} and \ex{innings} relate to a domain corresponding to ``sport'', thereby allowing the text domains to be learned automatically.

\label{sec:dis}
Several possible distributions can be used to model $\vect{z} \in \mathbb{R}^k$. Here we consider the following distributions:
\begin{description}
\item[Beta] which bounds all elements to the range $[0,1]$, such that $\vect{z}$ lies in a hyper-cube;
\item[Dirichlet] which also bounds all elements, as for Beta, however $\vect{z}$ are also constrained to lie in the probability simplex.
\end{description}
In both cases,\footnote{We also compared Gamma distributions, but they underperformed Beta and Dirichlet models.} each dimension of $\vect{z}$ is controlled by different distribution parameters, themselves formulated as different non-linear functions of $\xvect$.
We expect the Dirichlet model to perform the best, based on their widespread use in topic models, and their desirable property of generating a normalised vector, resembling common attention mechanisms \cite{bahdanau2014neural}.

Depending on the choice of distribution, the prior is modelled as
\begin{subequations} \label{eq:prior}
\begin{align}
&& p(\zvect | \xvect) =& 
\BetaD\left( \pmb{\alpha}^{B}, \pmb{\beta}^{B} \right)   \\
\mbox{or} && p(\zvect | \xvect) =& \Dirichlet\left( \alpha_0 \pmb{\alpha}^{D} \right)  \, , \label{eq:dir}
\end{align}
\end{subequations}
where the prior parameters are parameterised as neural networks of the input.
For the Beta prior, 
\begin{subequations} 
\begin{align}
 \pmb{\alpha}^{B} &= \D{elu} (\f_{\alpha,B}( \vect{x}) ) + 1\\
 \pmb{\beta}^{B} &= \D{elu} (\f_{\beta,B}( \vect{x}) ) + 1 \, ,
\end{align}
\end{subequations} 
where $\D{elu}(\cdot)+1$ is an element-wise activation function which returns a positive value \cite{clevert2015fast}, and $\f_\omega(\cdot)$ is a nonlinear function with parameters $\omega$---here we use a CNN.
The Dirichlet prior uses a different parameterisation,
\begin{subequations} 
\begin{align}
\alpha_0 =& \D{exp} (\f_{D,0} ( \vect{x} )) \\
\pmb{\alpha}^{D} =& \D{sigmoid}(\f_{D}( \vect{x})) \, , \label{eq:dirparams}
\end{align}
\end{subequations} 
where $\alpha_0$ is a positive-valued overall concentration parameter, used to scale all components in (\ref{eq:dir}), thus capturing overall sparsity, while $\pmb{\alpha}^{D}$ models the affinity to each channel.

\subsection{Variational Inference}
\label{sec:vi}
Using continuous latent variables, as described in \secref{sec:continuous}, gives rise to intractable inference; for this reason we develop a variational inference method based on the variational auto-encoder \cite{kingma2013auto}.
Fitting the model involves maximising the evidence lower bound (ELBO),
\begin{align} 
\log p_{\phi,\theta}(y|\xvect)  = & \log \int_{\zvect} p_\phi(\zvect \vert \xvect) p_\theta(y | \zvect, \xvect) \nonumber \\
 \geq & \expe_{q_\sigma} \left[ \log p_\theta(y | \zvect,\xvect) \right] \label{eq:elbo} \\
& - \lambda \dist_{\text{KL}} \big(q_{\sigma}(\zvect| \xvect,y,d) || p_\phi(\zvect \vert \xvect)\big) \, ,  \nonumber
\end{align}
where $q_\sigma$ is the variational distribution, parameterised by $\sigma$, chosen to match the family of the prior (Beta or Dirichlet) and $\lambda$ is a hyper-parameter controlling the weight of the KL term. 
The  ELBO in (\ref{eq:elbo}) is maximised with respect to $\sigma, \phi$ and $\theta$, using stochastic gradient ascent, where the expectation term is approximated using a single sample, $\hat{\vect{z}} \sim q_\sigma$, which is used to compute the likelihood directly.
Although it is not normally possible to backpropagate gradients through a sample, which is required to learn the variational parameters $\sigma$, this problem is usually side-stepped using a reparameterisation trick \cite{kingma2013auto}.
However this method only works for a limited range of distributions, most notably the Gaussian distribution, and for this reason we use the implicit reparameterisation gradient method \cite{DBLP:journals/corr/abs-1805-08498},
which allows for inference with a variety of continuous distributions, including Beta and Dirichlet.
We give more details of the implicit reparameterisation method in  \appref{app:irg}.

The variational distribution $q$, is defined in an analagous way to the prior, $p$, see  (\ref{eq:prior}--\ref{eq:dirparams}), i.e., using a neural network parameterisation for the distribution parameters.
The key difference is that $q$  conditions not only on $\xvect$ but also on the target label $y$ and domain $d$.
This is done by embedding both $y$ and $d$, which are concatenated with a CNN encoding of $\xvect$, and then transformed into the distribution parameters. 
Semi-supervised learning with respect to the domain can easily be facilitated by setting $d$ to the domain identifier when it is observed, otherwise using a sentinel value $d=\texttt{UNK}$, for domain-unsupervised instances.
The same trick is used for $y$, to allow for vanilla semi-supervised learning (with respect to target label).
The use of $y$ and $d$ allows the inference network to learn to encode these two key variables into $z$, to encourage the latent variable, and thus model channels, to be informative of both the target label and the domain.
This, in concert with the KL term in (\ref{eq:elbo}), ensures that the prior, $p$, must also learn to discriminate for domain and label, based solely on the input text, $\xvect$. 

For inference at test time, we assume that only $\xvect$ is available as input, and accordingly the inference network cannot be used.
Instead we generate a sample from the prior $\hat{\vect{z}} \sim p(\vect{z}|\xvect)$, which is then used to compute the maximum likelihood label, $\hat{y} = \operatorname{arg\,max}_y p(y|\xvect,\hat{\vect{z}})$.
We also experimented with Monte Carlo methods for test inference, in order to reduce sampling variance, using:
(a) prior mean $\bar{\vect{z}} = \mu$; (b) Monte Carlo averaging $\bar{y} = \frac{1}{m} \sum_{i} p(y|\xvect,\hat{\vect{z}}_i)$ using $m=100$ samples from the prior; and (c) importance sampling \cite{glynn1989importance} to estimate $p(y|\xvect)$ based on sampling from the inference network, $q$.%
\footnote{Importance sampling estimates $p(y|\xvect) = \expe_q[p(y,\vect{z}|\xvect)/ q(\vect{z}|\xvect, y, d) ]$ for each setting of $y$ using $m=100$ samples from $q$, and then finds the maximising $y$. This is tractable in our settings as $y$ is a discrete variable, e.g., a binary sentiment, or multiclass language label. }
None of the Monte Carlo methods showed a significant difference in predictive performance versus the single sample technique, although they did show a very tiny reduction in variance over 10 runs. 
This is despite their being orders of magnitude slower, and therefore we use a single sample for test inference hereafter.

\section{Experiments}

\subsection{Multi-domain Sentiment Analysis}
To evaluate the proposed models, we first experiment with a multi-domain sentiment analysis dataset, focusing on out-of-domain evaluation where the test domain is unknown.

\begin{table}
\centering
\footnotesize
\begin{tabular}{lcc}
\toprule
Domain & $\mathcalA(\xvect, y, d)$\!\! & $\mathcalC(\xvect, y, ?)$ \\
\midrule
apparel
  & 1,000  & 1,000\\
baby
  & 950  & 950 \\
camera \& photo
  & 1000 & 999 \\
health \& personal care \!\!\!\!\!\!\!\!
  & 1,000 & 1,000 \\
magazines
  & 985 & 985 \\
music
  & 1,000 & 1,000 \\
sports \& outdoors
  & 1,000 & 1,000 \\
toys \& games
  & 1,000 & 1,000 \\
video
  & 1,000 & 1,000 \\
\bottomrule
\end{tabular}%

  \caption{Numbers of instances (reviews) for each training domain in our dataset, under
    the two categories $\mathcalA$ (domain and label known) and $\mathcalC$
    (label known; domain unknown), in which ``$?$'' represents the
    ``\resource{unk}'' token, meaning the given attribute is
    unobserved.}
\label{tab:dat:pr}
\end{table}

\begin{table*}[t]
  \sisetup{detect-weight=true,detect-inline-weight=math}
  \centering
\footnotesize
  \begin{tabular}{lll *{5}{ S[table-format=2.1,round-mode=places,round-precision=1]@{~$\pm$~}S[table-format=1.1,round-mode=places,round-precision=1] }}

\toprule
Data & & & \multicolumn{2}{c}{\resource{B}} & \multicolumn{2}{c}{\resource{D}} & \multicolumn{2}{c}{\resource{E}} & \multicolumn{2}{c}{\resource{K}} & \multicolumn{2}{c}{Average} \\
\midrule




 

\multirow{8}{*}{$\mathcalA+\mathcalC$}
& \scnn &
 & 78.86836  & 1.3081189515
 & 80.900111 & 1.519653083
 & 82.389832 & 0.8320603276    
 & 84.087102 & 1.847597894
 & 81.56135125 & 0.8654609782 \\

& \bcnn &
 & 78.98  & 1.50
 & 82.5 & 1.29
 & 84.1 & 0.8
 & 85.9 & 0.8
 & 82.875 & 0.9 \\

& \gen &
 & 78.415283 & 0.9191596196
 & 81.228758 & 0.9907181187
 & 83.908046 & 1.72621948   
 & \bfseries 87.542088 & 1.216097799
  & 82.77354375 & 1.112017965 \\

\cmidrule{2-13}
& \dsda &
 & 76.782753 & 1.38370508
 & 79.647436 &  1.69386401
 & 83.102564 & 1.48752788
 & 85.761048 & 1.95872286
 & 81.32345025 & 1.04568417 \\
& & + semi-sup. 
 & 77.057356 & 1.567951647
 & 79.891566 & 0.9981150274
 & 83.089976 & 1.709384462
 & 85.438422 & 1.329364947
 & 81.36933 & 0.642114293\\

\cmidrule{2-13}

& \csda
 & w. $\BetaD$
 & 78.443116 & 0.8242547713
 & \bfseries 84.38735 & 0.7343856566
 & 82.907665 & 1.05950916
 & 87.184466 & 1.331193892
 & 83.23064925 & 0.8748912608 \\
 
& & w. $\Dirichlet$
 & \bfseries 79.96071 & 1.372028392
 & 84.320633 & 1.352330438
 & \bfseries 86.206897 & 1.459021358
 & 87.037039 & 0.2673026358
 & \bfseries 84.38131975 & 0.9030758728 \\

\toprule

\multirow{8}{*}{$\mathcalA$ only}
& \scnn &
  & 76.040922 & 1.79822004
  & 76.996881 & 0.9744379
  & 81.496703 & 1.27250065
  & 82.791497 & 1.61028685
  & 79.331501 & 0.71347525 \\

& \bcnn &
 & 76.7 & 1.8
 & 79.2 & 0.4
 & 82.01 & 1.2
 & 83.1 & 1.8
 & 79.752 & 1.3 \\

& \gen &
  & 76.70444862 & 1.97440116
  & 79.120229   & 1.31680398
  & 82.09291338 & 1.58899696
  & 83.97756375 & 1.14427069
  & 80.47378858 & 0.7136388 \\

\cmidrule{2-13}
& \dsda &
  & 74.34002863 & 1.38897789
  & 75.768819   & 2.24461255
  & 80.49237725 & 1.34421836
  & 82.82360712 & 1.36867576
  & 78.35620788 & 0.93619439 \\

& & + unsup. 
  & 74.1301485  & 1.98869291
  & 75.6203881  & 2.25496525
  & 80.7817275  & 1.30520554
  & 82.9684779  & 1.73425466
  & 78.3751855  & 0.59757994 \\

\cmidrule{2-13}

& \csda & w. $\BetaD$
  & \bfseries 78.03722557 & 1.93627393
  & 80.45020757 & 1.05809562
  & 83.69041371 & 1.25104212
  & 85.66479171 & 1.31314603
  & 81.96065973 & 1.13493416 \\
  
& & w. $\Dirichlet$
   & 77.94928786 & 1.64482842
   & \bfseries 80.597083 & 0.87988201
   & \bfseries 84.36105029 & 1.129751
   & \bfseries 86.45998471 & 0.8507671
   & \bfseries 82.34185145 & 0.56285258 \\

\midrule
& \multicolumn{2}{l}{\resource{in domain}$\clubsuit$}
  & \multicolumn{2}{l}{80.4} & \multicolumn{2}{l}{82.4} & \multicolumn{2}{l}{84.4} &  \multicolumn{2}{l}{87.7}
  & \multicolumn{2}{l}{83.7} \\

\bottomrule

\end{tabular}%

\caption{Accuracy [\%] and standard deviation of different models under
  two data configurations: (1) using both $\mathcalA$ and $\mathcalC$
(domain semi-supervised learning); and (2) using $\mathcalA$ only
(domain supervised learning). In each case, we evaluate over the four
held-out test domains (\resource{B}, \resource{D}, \resource{E} and \resource{K}),
and also report the accuracy.
Best results are indicated in \textbf{bold} in each configuration.
Key: $\clubsuit$ from \newcite{blitzer2007biographies}.}
\label{tab:res:pr}
\end{table*}

We derive our dataset from Multi-Domain Sentiment Dataset v2.0 \cite{blitzer2007biographies}.\footnote{From \url{https://www.cs.jhu.edu/~mdredze/datasets/sentiment/}.}
The task is to predict a binary sentiment label, i.e., positive vs.\ negative.
The unprocessed dataset has more than 20 domains.
For our purposes, we filter out domains with fewer than 1k labelled instances or fewer than 2k unlabelled instances, resulting in 13 domains in total.

To simulate the semi-supervised domain situation,
we remove the domain attributions for one half of the labelled data, denoting them as domain-unlabelled data $\mathcalC(\xvect,y,?)$.
The other half are sentiment- and domain-labelled data $\mathcalA(\xvect,y,d)$.
%
We present a breakdown of the dataset in \tabref{tab:dat:pr}.\footnote{The dataset, along with the source code, can be found at \url{https://github.com/lrank/Code_VariationalInference-Multidomain}}

For evaluation, we hold out four domains---namely books
(``\resource{B}''), dvds (``\resource{D}''), electronics
(``\resource{E}''), and kitchen \& housewares (``\resource{K}'')---for comparability with previous work \cite{blitzer2007biographies}.
Each domain has 1k test instances, and
we split this data into \emph{dev} and \emph{test} with ratio 4:6.
The \emph{dev} dataset is used for hyper-parameter tuning and early stopping,\footnote{This confers light supervision in the target domain. However we would expect similar results were we to use disjoint held out domains for development wrt testing.}
and we report accuracy results on \emph{test}.

\subsubsection{Baselines and Comparisons}

For comparison, we use 3 baselines.
The first is a single channel $\CNN$ (``\scnn''), which jointly over all data instances in a single model, without domain-specific parameters.
The second baseline is a multi channel $\CNN$ (``\bcnn''), which expands the capacity of the \scnn model (606k parameters) to match \csda and \dsda (roughly 7.5m-8.3m parameters).
Our third baseline is a multi-domain learning approach
using adversarial learning for domain generation (``\gen''), the 
best-performing model of \newcite{yitong2018naacl} and state-of-the-art
for unsupervised multi-domain adaptation over a comparable
dataset.\footnote{The dataset used in \newcite{yitong2018naacl} differs
  slightly in that it is also based off Multi-Domain Sentiment Dataset
  v2.0, but uses slightly more training domains and a slightly different
  composition of training data. We retrain the model of the authors over
  our dataset, using their implementation.}
We report results for their best performing \gen\texttt{+d+g} model.

\subsubsection{Training Strategy}
\label{ssec:train}
For the hyper-parameter setups, we provide the details in \appref{app:ba}.
In terms of training, we simulate two scenarios using two experimental
configurations, as discussed above: (a) domain supervision; and
(2) domain semi-supervision.  For domain supervised training, only
$\mathcalA$ is used, which covers only 9 of the domains, and 
the test domain data is entirely unseen.
For domain semi-supervised training, we use combinations of $\mathcalA$ 
and $\mathcalC$, noting that both sub-corpora
do not include data from the target domains, and none of which is
explicitly labelled with sentiment, $y$, and domain, $d$. 
These simulate the setting where we have heterogenous data which includes a lot of 
relevant data, however its metadata is inconsistent, and thus cannot be easily modelled.

For $\lambda$ in (\ref{eq:elbo}), according to the derivation of the ELBO it should be the case that $\lambda=1$, however other settings are often justified in practice \cite{alemi2018fixing}.  
Accordingly, we tried both annealing and fixed schedules, but found no consistent differences in end performance.
We performed a grid search for the fixed value, $\lambda = 10^a ,\,  a \in \{ -3,  -2, -1, 0, 1\}$, and selected $\lambda=10^{-1}$, based on development performance.
We provide further analysis in the form of a sensitivity plot in \secref{sec:lbd}.
The latent domain size $k$ for \dsda is set to the true number of training domains $k=D=9$.
Note that, even for \dsda, we could use  $k \ne D$, which we explore in
the $\mathcalA+\mathcalC$ supervision setting in \secref{sec:results}.
For \csda we present the main results with $k=13$, set to match the total number of domains in training and testing.

\subsubsection{Results}
\label{sec:results}

\tabref{tab:res:pr} reports the performance of
different models under two training configurations: (1) with
$\mathcalA+\mathcalC$ (domain semi-supervised learning);  and (2)
with $\mathcalA$ only (domain supervised learning). In each
case, we report the standard deviation based on 10 runs with different
random seeds.

Overall, domain \resource{B} and \resource{D} are more difficult than
\resource{E} and \resource{K}, consistent with previous work.
Comparing the two configurations, we see that when we use domain semi-supervised training (with the addition of $\mathcalC$), all models perform better, demonstrating the utility of domain semi-supervised learning when annotated data is limited.

Comparing our discrete and continuous approaches (\dsda and \dsda,
resp.), we see that \csda consistently performs the best, outperforming the baselines
by a substantial margin.
In contrast \dsda is disappointing, underperforming the baselines, and moreover, shows no change in performance between domain  supervision versus the semi-supervised or unsupervised settings. 
Among the \csda based methods, all the distributions perform well,
but the Dirichlet distribution performs the best overall, which we attribute to better
modelling of the sparsity of domains, thus reducing the influence of uncertain 
and mixed domains.
The best results are for domain semi-supervised learning ($\mathcalA+\mathcalC$), which brings an increase in accuracy of about 2\% over domain supervised learning ($\mathcalA$) consistently across the different types of model.

\subsection{Analysis and Discussion}
To better understand what the model learns,
we focus on the \csda model, using the Dirichlet distribution.

\begin{figure}[t]
\resizebox{\columnwidth}{!}{\input{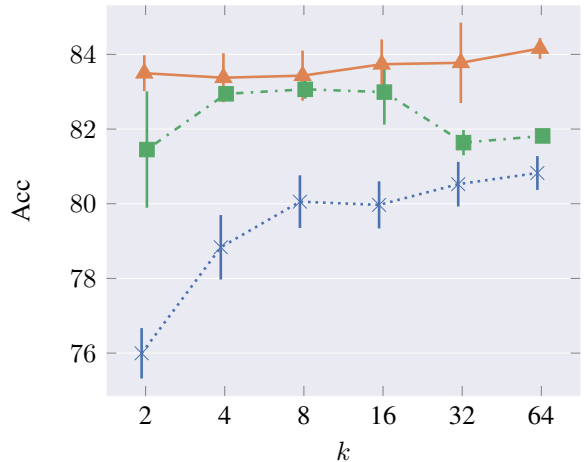}}
\caption{Performance with standard error ($\pmb{|}$) as latent domain size $k$ is increased in
  $\log 2$ space with \dsda (\ref{fig:mark:b}) and with three \csda methods using $\BetaD$ (\ref{fig:mark:e}) and $\Dirichlet$
  (\ref{fig:mark:d}) averaged accuracy, over $\mathcalA+\mathcalC$.}
\label{fig:lat}
\end{figure}

First, we consider the model capacity, in terms of the latent domain
size, $k$.
\figref{fig:lat} shows the impact of varying $k$. 
Note that the true number of domains is $D=13$, comprising 9 training and 4 test domains. 
Setting $k$ to roughly this value appears to be justified, in that the mean accuracy increases with $k$, and plateaus around $k = 16$. 
Interestingly, when $k \geq 32$, the performance of \csda with $\BetaD$ drops, while performance for $\Dirichlet$ remains high---indeed $\Dirichlet$ is consistently superior even at the extreme value of $k=2$, although it does show improvement as $k$ increases.
Also observe that \dsda requires a large latent state inventory, supporting our argument for the efficiency of continuous cf.\@ discrete latent variables.

\begin{figure*}[t]
\includegraphics[width=\textwidth]{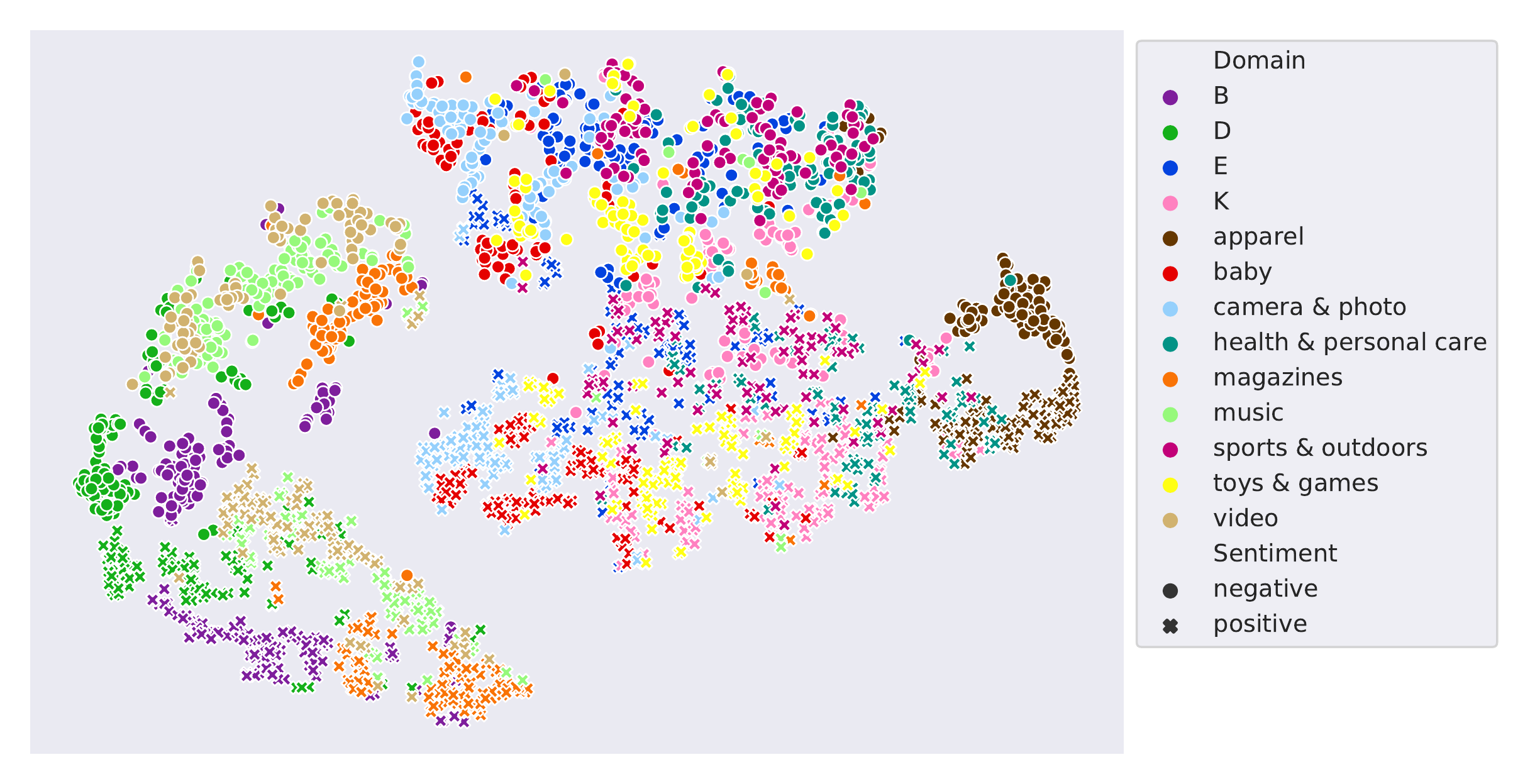}
\caption{t-SNE of hidden representations in \csda over all 13 domains, comprising 4 held-out testing domains (B, D, E and K), and the remaindering 9 domains are used only for training. 
Each point is a document, and the symbol indicates its gold sentiment label, using a filled circle for negative instances and cross for positive.
}
\label{fig:tsne}
\end{figure*}

\begin{table}[t]
  \sisetup{detect-weight=true,detect-inline-weight=math}
  \centering
\footnotesize
\begin{tabular}{l *{5}{ S[table-format=2.1,round-mode=places,round-precision=1] }}
\toprule
\csda & {\resource{B}} & {\resource{D}} & {\resource{E}} & {\resource{K}} & {Average} \\
\midrule
$\mathcalA$ & 77.94928786    & 80.597083    & 84.36105029   & 86.45998471   & 82.34185145 \\
$\mathcalA+\mathcalC$ & 79.96071 & 84.320633 & 86.206897 & 87.037039 & 84.38131975 \\
$\mathcalC$ & 77.5595304 & 81.5389806 & 83.7492892 & 85.2290837 & 82.01922098 \\
\bottomrule
\end{tabular}
\caption{Accuracy [\%] of \csda w. Dirichlet trained with different configurations of $\mathcalA$ and $\mathcalC$.}
\label{tab:res:b}
\end{table}

Next, we consider the impact of using different combinations of $\mathcalA$ and $\mathcalC$.
\tabref{tab:res:b} shows the performance of difference configurations.
Overall, $\mathcalA+\mathcalC$ gives excellent performance. 
Interestingly, $\mathcalC$ on its own is only a little worse than only $\mathcalA$, showing that target labels $y$ are more important for learning than the domain $d$. 
The $\mathcalC$ configuration fully domain unsupervised training still results in decent performance, boding well for application to very messy and heterogenous datasets with no domain metadata.

\begin{figure}[t]
\resizebox{\columnwidth}{!}{\begin{tikzpicture}
\begin{axis}[
    xmode=log,
    log ticks with fixed point,
    xlabel={$\lambda$},
    ylabel={Acc},
    grid=major,
    legend entries={$y$,$d$},
]
\addplot table {
1e-3  97.31543624 
3e-3  86.10639249 
1e-2  93.51464435 
3e-2  90.02079002 
1e-1  80.15481768 
3e-1  80.77451758 
1  76.88274547 
3e0  80.82924309 
1e1  79.528403 
};
\addplot table {
1e-3 25.53211889
3e-3 31.04604381
1e-2 23.56171548
3e-2 17.72349272
1e-1 17.54471379
3e-1 17.71081153
1 18.14585319
3e0 16.7398605
1e1 13.37084673
};
\draw[blue, dashed] (axis cs:1e-3, 50) -- (axis cs:10, 50);

\draw[red, dashed] (axis cs:1e-3, 11.11) -- (axis cs:10, 11.11);

\end{axis}
\end{tikzpicture}}
\caption{Diagostic classifier accuracy [\%] over $\zvect$ to predict the sentiment label $y$ and domain label $d$, with respect to different $\lambda$, shown on a log scale. Dashed horizontal lines show chance accuracy for both outputs.}
\label{fig:diag}
\end{figure}
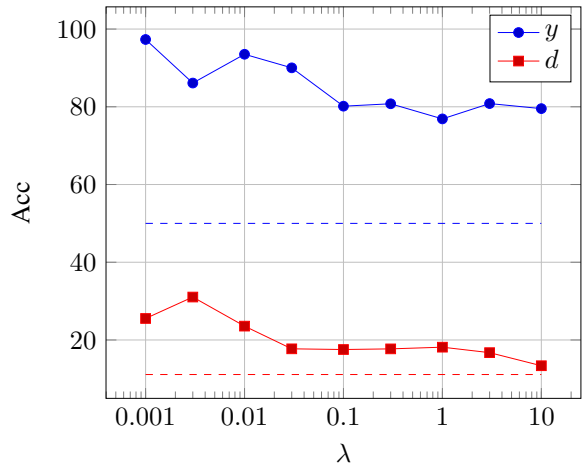

Finally, we consider what is being learned by the model, in terms of how it learns to use the $k$ dimensional latent variables for different types of data. 
We visualise the learned representations, showing points for each domain plotted in a 2d t-SNE
plot \cite{maaten2008visualizing} in \figref{fig:tsne}.
Notice that each domain is split into two clusters, representing positive
($\pmb{\times}$) and negative (\textbullet) instances within that
domain.  Among the test domains,
\textcolor[HTML]{\Bcolour}{\resource{B}} (books) and
\textcolor[HTML]{\Dcolour}{\resource{D}} (dvds) are clustered close
together but are still clearly separated, which is encouraging given the close
relation between these two media.  The other two,
\textcolor[HTML]{\Ecolour}{\resource{E}} (electronics) and
\textcolor[HTML]{\Kcolour}{\resource{K}} (kitchen \& housewares) are
mixed together and intermingled with other domains.  Overall across all
domains, the \textcolor[HTML]{653700}{\resource{apparel}} cluster is
quite distinct, while \textcolor[HTML]{d1b26f}{\resource{video}} and
\textcolor[HTML]{96f97b}{\resource{music}} are highly associated with
\textcolor[HTML]{\Dcolour}{\resource{D}}, and part of the cluster for
\textcolor[HTML]{f97306}{\resource{magazines}} is close to
\textcolor[HTML]{7e1e9c}{\resource{B}}; all of these make sense
intuitively, given similarities between the respective products.
\textcolor[HTML]{\Ecolour}{\resource{E}} is related to
\textcolor[HTML]{95d0fc}{\resource{camera}} and
\textcolor[HTML]{dede14}{\resource{games}}, while
\textcolor[HTML]{\Kcolour}{\resource{K}} is most closely connected to
\textcolor[HTML]{029386}{\resource{health}} and
\textcolor[HTML]{c20078}{\resource{sports}}.

\begin{table*}[t]
  \sisetup{detect-weight=true,detect-inline-weight=math}
  \centering
\footnotesize
\begin{tabular}{lll *{8}{ S[table-format=2.1,round-mode=places,round-precision=1] }}

\toprule
\multicolumn{2}{l}{Data} & & \resource{EuroGov} & \resource{TCL} & \resource{Wikipedia} & \resource{EMEA} & \resource{EuroPARL} & \resource{TBE} & \resource{TSC} & {Average} \\
\midrule
\parbox[t]{3mm}{\multirow{5}{*}{ $\mathcalC$ }}
& \scnn &
  & 98.7555554  & 92.2591944  & 85.851802 & 98.499166 & 92.3291004  & 79.3308826  & 91.7348066  & 91.25150105714287 \\
& \bcnn &
 & \bfseries 98.8740742  & \bfseries 93.5831874  & 86.223416 & 99.16398  & 95.9767196  & 88.3333334  & 91.7348066  & 93.41278817142856 \\
\cmidrule{2-11}
& \dsda &
 & 98.325926 & 91.887916 & 86.317439 & 97.807671 & 95.1650794  & 85.9754902  & 79.0202576  & 90.6428256 \\
& \csda & w. $\BetaD$
 & 98.7185184  & 93.0462346  & \bfseries 88.9963286  & \bfseries 99.2540244  & 96.7544974  & 93.0867646  & \bfseries 95.2169428 & \bfseries 95.00904440 \\
& & w. $\Dirichlet$
 & \bfseries 98.8888886  & 93.0290718  & \bfseries 89.0366242  & 99.1617564  & \bfseries 96.6878306  & \bfseries 93.235294 & 94.4950276  & 94.93349902857145 \\

\midrule

\parbox[t]{3mm}{\multirow{3}{*}{ $\mathcalA$ }}
& \dsda &
  & 98.0  & 91.754816 & 85.6861428  & 97.6987214  & 95.332275 & 85.3627454  & 78.062615 & 90.27104508571429 \\
& \csda & w. $\BetaD$
  & \bfseries 99.3037038  & \bfseries 93.7022764  & 89.1336466  & 99.2440358  & \bfseries 96.9492064  & \bfseries 93.6299018  & 93.8710866  & 95.12197962857142 \\
& & w. $\Dirichlet$
   & 98.9629628  & \bfseries 93.6882662  & \bfseries 89.3037832  & \bfseries 99.3029458  & \bfseries 96.889947 & 93.259804 & \bfseries 96.1104972  & \bfseries 95.35974374285713 \\

\midrule
& \multicolumn{2}{l}{\gen}
&  99.9 & 93.1 & 88.7 & 92.5 & 97.1 & 91.2 & 96.1 & 94.08 \\
& \multicolumn{2}{l}{\resource{langid.py}}
&  98.7 & 90.4 & 91.3 & 93.4 & 97.4 & 94.1 & 92.7 & 94.0 \\

\bottomrule
\end{tabular}%


\caption{Accuracy [\%] over 7 LangID benchmarks, as well as the averaged score, for different models under
  two data configurations: (1) using domain unsupervised learning ($\mathcalC$); and (2) using domain supervised learning ($\mathcalA$). 
The best results are indicated in \textbf{bold} in each
configuration. Note that the training data for \gen and \resource{langid.py} is slightly different from that used in the original papers.}
\label{tab:res:lid}
\end{table*}

\label{sec:lbd}

To obtain a better understanding of what is being encoded in the latent variable, and how this is effected by the setting of $\lambda$, we learn simple diagnostic classifiers  to predict sentiment label $y$ and domain label $d$, given only $\zvect$ as input.
To do so, we first train our model over the training set, and record samples of $\zvect$ from the inference network.
We then partition the training set, using 70\% to learn linear logistic regression classifiers to predict $y$ and $d$, and use the remaining 30\% for evaluation.
\figref{fig:diag} shows the prediction accuracy, based on averaging over three runs, each with different $\zvect$ samples.
Clearly very small $\lambda \le 10^{-2}$, leads to almost perfect sentiment label accuracy which is evidence of overfitting by using the latent variable to encode the response variable. 
For $\lambda \ge 10^{-1}$ the sentiment accuracy is still above chance, as expected, but is more stable.
For the domain label $d$, the predictive accuracy is also above chance, albeit to a lesser extent, and shows a similar downward trend.
At the setting $\lambda = 0.1$, used in the earlier experiments, this shows that the latent variable encodes captures substantial sentiment, and some domain knowledge, as observed in Figure~\ref{fig:tsne}.

In terms of the time required for training, a single epoch of training took about 25min for the \csda method, using the default settings, and a similar time for \dsda and \bcnn. The runtime increases sub-linearly with increasing latent size $k$.

\subsection{Language Identification}
To further demonstrate our approaches, we then evaluate our models with the second task, language identification (LangID: \citet{Jauhiainen+:2018}).

For data processing, we use 5 training sets from 5 different domains with 97 language, following the setup of \citet{DBLP:conf/ijcnlp/LuiB11}.
We evaluate accuracy over 7 holdout benchmarks: \resource{EuroGov}, \resource{TCL}, \resource{Wikipedia} from \citet{DBLP:conf/naacl/BaldwinL10}, \resource{EMEA} \cite{tiedemann2009news}, \resource{EuroPARL} \cite{koehn2005europarl}, \resource{TBE} \cite{tromp2011graph} and \resource{TSC} \cite{carter2013microblog}.
Differently from sentiment tasks, here, we evaluate our methods using the full dataset, but with two configurations: (1) domain unsupervised, where all instance have only labels but no domain (denoted $\mathcalC$); and (2) domain supervised learning, where all instances have labels and domain ($\mathcalA$).

\subsubsection{Results}
\tabref{tab:res:lid} shows the performance of different models over 7 holdout benchmarks and the averaged scores.
We also report the results of \gen, the best model from~\citet{yitong2018naacl}, and one state-of-the-art off-the-shelf LangID tool: \resource{langid.py}~\cite{lui2012langid}.
Note that, both \scnn and \bcnn are domain unsupervised methods.
In terms of results, overall, both of our \csda models consistently outperform all other baseline models.
Comparing the different \csda variants, $\BetaD$ vs.~$\Dirichlet$, both perform closely across the LangID tasks.
Furthermore, \csda out-performs the state-of-the-art in terms of average scores.
Interestingly the two training configurations show that domain knowledge
$\mathcalA$ provides a small performance boost for \csda, but not does help for \dsda.
Above all, the LangID results confirm the effectiveness of our proposed approaches.

\section{Related Work}

Domain adaptation (``DA'') typically involves one or
more training domains and a single target domain.  Among DA approaches,
single-domain adaptation is the most common scenario, where a model is
trained over one domain and then transferred to a single target domain
using prior knowledge of the target domain
\cite{blitzer2007biographies,glorot2011domain}.  
Adversarial learning methods have been proposed for learning robust domain-independent
representations, which can capture domain knowledge through semi-supervised
learning \cite{ganin2016domain}.

Multi-domain adaptation uses training data from more than one training domain. 
Approaches include feature augmentation methods
\cite{daume2007frustratingly}, and  analagous neural models \cite{Joshi2012multi,C16-1038}, 
as well as attention-based and hierarchical methods \cite{li2018hierarchical}.  
These works assume the `oracle' source domain is known when transferring, however we do not require an oracle in this paper.
Adversarial training methods have been employed to learn robust
domain-generalised representations \cite{DBLP:conf/emnlp/LiuQH16}.
\newcite{yitong2018naacl} considered the case of the model having
no access to the target domain, and using adversarial learning to generate
domain-generation representations by cross-comparison between source domains.

The other important component of this work is Variational Inference
(``VI''), a method from machine learning that approximates probability
densities through optimisation
\cite{blei2017variational,kucukelbir2017automatic}.  The idea of a
variational auto-encoder has been applied to language generation \cite{bowman2015generating,kim2018semi,miao2017discovering,zhou2017multi,
  zhang2016variational} and machine translation \cite{shah2018generative,eikema2018auto}, but not in the context of semi-supervised domain adaptation.

\section{Conclusion}
In this paper, we have proposed two models---\dsda and \csda---for
multi-domain learning, which use a graphical model with a latent variable to 
represent the domain.
We propose models with a discrete latent variable, and a continuous vector-valued latent variable,
which we model with Beta or Dirichlet priors.
For training, we adopt a variational inference technique based on the variational autoencoder.  
In empirical evaluation over a multi-domain sentiment dataset and seven language identification benchmarks, our models outperform strong baselines, across varying data conditions, including a setting where no target domain data is provided.  
Our proposed models have broad utility across NLP applications on heterogenous corpora.

\section*{Acknowledgements}
This work was supported by an Amazon Research Award.
We thank the anonymous reviewers for their helpful feedback and suggestions.

\bibliographystyle{acl_natbib}
\bibliography{ref}

\clearpage

\appendix

\section{Appendices}
\subsection{Base Model Architecture}
\label{app:ba}

For the sentiment task, all the hidden representations
are learned by convolutional neural networks ($\CNN$), following \citet{kim2014convolutional}.
All documents are lower-cased and truncated to maximum 256
tokens, and then each word is mapped into a 300 dimensional vector representation using
randomly-initialised word embeddings.
In each $\CNN$ channel, filter windows are set to $\{3, 4, 5\}$, with 128 filters for each.
Then, $\D{ReLU}$ and $\D{pooling}$ are applied after the filtering, generating $384$-d ($128 * 3$) hidden representations.
Dropout is applied to the hidden $\hvect$, at a rate of 0.5.
For simplicity, we use the same $\CNN$ architecture to encode the functions $\f$ used in the prior $q$ and in the inference networks $p$, in each case with different parameters.
Specifically, in prior $q$, the embedding sizes of domain and label are set to 16 and 4, respectively.
$\pmb{\alpha}$ and $\pmb{\beta}$ share the same $\CNN$ but with different output projections.
After gating using $\zvect$, the final hidden goes through a one-hidden \MLP with hidden size $300$.
We use the Adam optimiser \cite{kingma2014adam} throughout, with the learning
rate set to $10^{-4}$ and a batch size of 32, optimising the loss functions
(\ref{eq:dsdaloss}) or (\ref{eq:elbo}), for \dsda and \csda, respectively.

For the language identification task, all documents are tokenized as a byte sequence, truncated or padded to a length of 1k bytes.
We use the same $\CNN$ architecture and hyper-parameter configurations as for the sentiment task.

\subsection{Implicit Reparameterisation Gradient}
\label{app:irg}

In this section, we outline the  implicit reparameterisation gradient method of \newcite{DBLP:journals/corr/abs-1805-08498}.
\begin{subequations}
First, we review some background on variational inference.
We start by defining a differentiable and invertible standardization function as
\begin{align}\label{equ:irg_def}
S_\sigma(\zvect) = \epsilon \sim q(\epsilon) \, ,
\end{align}
which describes a mapping between points drawn from a specific distribution function and a standard distribution, $q$.
For example, for a Gaussian distribution $z \sim \mathcal{N}(\mu,\psi)$, we can define $S_{\mu,\psi}(z) = (z - \mu)/\psi \sim \mathcal{N}(0,1)$ to map to the standard Normal.
We aim to compute the gradient of the expectation of a objective function $f(\zvect)$, 
\begin{align}
\nabla_\sigma \expe_{q_\sigma(\zvect)}[f(\zvect)] = \expe_{q(\epsilon)} [\nabla_{\sigma}f(S^{-1}(\epsilon) ) ] \, ,
\end{align}
where in ELBO (\ref{eq:elbo}) in our case, $f(\zvect) = p_\theta(y | \zvect,\xvect)$ is the likelihood function. 

The implicit reparameterisation gradient technique is a way of computing the reparameterisation without the need for inversion of the standardization function.
This works by applying $\nabla_{\sigma}S^{-1}(\epsilon) = \nabla_\sigma \zvect$,
\begin{align} \label{equ:irg_fin}
\nabla_\sigma \expe_{q_\sigma(\zvect)}[f(\zvect)] = \expe_{q_\sigma(\zvect)} [\nabla_{\zvect}f(\zvect) \nabla_{\sigma}\zvect ] \, .
\end{align}
However, we still need to calculate $\nabla_\sigma \zvect$.
The key insight here is that we can compute $\nabla_\sigma \zvect$ by \emph{implicit differentiation}.
We apply the total gradient $\nabla^{\text{TD}}_{\sigma}$ over (\ref{equ:irg_def}),
\begin{align} \label{equ:irg_tg}
\nabla^{\text{TD}}_\sigma S_\sigma(\zvect) = \nabla^{\text{TD}}_\sigma \epsilon \, .
\end{align}
From the definition of a standardization function, the noise $\epsilon$ is independent of $\sigma$, and we apply the multi-variable chain rule over left side of (\ref{equ:irg_tg}),
\begin{align}
\frac{\partial S_\sigma(\zvect)}{\partial \zvect} \nabla_\sigma \zvect + \frac{\partial S_\sigma(\zvect)}{\partial \sigma} = \mathbf{0} \, .
\end{align}
Therefore, the key of the implicit gradient calculation in this process can be summarised as
\begin{align}
\nabla_\sigma \zvect = -(\nabla_\zvect S_\sigma(\zvect) )^{-1} \nabla_\sigma S_\sigma (\zvect) \, .
\end{align}
This expression allows for computation of (\ref{equ:irg_fin}), which can be applied to a range of distribution families.
We refer the reader to~\newcite{DBLP:journals/corr/abs-1805-08498} for further details.
\end{subequations}

\end{document}